\documentclass[11pt,a4paper]{article}

\usepackage[hyperref]{emnlp2020}  
\usepackage{times}    
\usepackage{latexsym}

\usepackage{microtype}

\aclfinalcopy 

\usepackage[utf8]{inputenc} 
\usepackage[T1]{fontenc}    
\usepackage{url}            
\usepackage{booktabs}       
\usepackage{amsfonts}       
\usepackage{amsmath}
\usepackage{nicefrac}       
\usepackage{graphicx}
\usepackage{microtype}      
\usepackage{tabularx}
\usepackage{xcolor}
\usepackage{bbm}
\usepackage{array}
\usepackage{arydshln}
\usepackage{amsfonts}
\usepackage{amsmath}

\usepackage{bbm}
\usepackage{boldline}
\usepackage{bigstrut}
\usepackage{blindtext}
\usepackage{booktabs, siunitx}
\usepackage[labelfont=bf, format=plain, justification=justified, singlelinecheck=false]{caption}
\usepackage{color}
\usepackage{cprotect}
\usepackage{ctable}
\usepackage{dirtytalk}
\usepackage{enumitem}
\usepackage[export]{adjustbox}
\usepackage{float}
\usepackage{graphicx}
\usepackage{hhline}
\usepackage{latexsym}
\usepackage{mathrsfs}
\usepackage{microtype}
\usepackage{moresize}
\usepackage{multicol}
\usepackage{multirow}
\usepackage{nccmath}
\usepackage{nicefrac}
\usepackage{pifont}
\usepackage{placeins}
\usepackage{soul}
\usepackage{subcaption}
\usepackage{times}
\usepackage[utf8]{inputenc}
\usepackage{url}
\usepackage{verbatim}
\usepackage{wrapfig, lipsum}
\usepackage{textcomp}
\usepackage{enumitem}

\newcommand\sL{\ensuremath{\mathcal{L}}}
\newcommand\sD{\ensuremath{\mathcal{D}}}

\definecolor{lblue}{HTML}{A6CEE3}
\definecolor{lgreen}{HTML}{B2DF8A}
\definecolor{lred}{HTML}{FB9A99}
\definecolor{lorange}{HTML}{FDBF6F}
\definecolor{mblue}{HTML}{80B1D3}
\definecolor{mgreen}{HTML}{B3DE69}
\definecolor{mred}{HTML}{FB8072}
\definecolor{morange}{HTML}{FDB462}
\definecolor{blue}{HTML}{1F78B4}
\definecolor{green}{HTML}{33A02C}
\definecolor{red}{HTML}{E31A1C}
\definecolor{orange}{HTML}{FF7F00}
\definecolor{dblue}{HTML}{0050EF}
\definecolor{dgreen}{HTML}{006D2C}
\definecolor{dorange}{HTML}{EC7014}

\newcommand{\dblue}[1]{{\color{dblue} #1}}

\newcommand{\dorange}[1]{{\color{dorange} #1}}

\newcommand{\cut}[1]{}
\newcommand{\xhdr}[1]{\noindent{\bfseries #1}.}

\title{Stronger Transformers for Neural Multi-Hop Question Generation}

\author{Devendra Singh Sachan$^{1,2}$, Lingfei Wu$^{3}$, Mrinmaya Sachan$^{4}$, William Hamilton$^{1,2}$ \\
$^{1}$Mila - Quebec AI Institute\\
$^{2}$School of Computer Science, McGill University\\
$^{3}$IBM Thomas J. Watson Research Center, Yorktown Heights\\
$^{4}$ETH Zurich\\
{\tt sachande@mila.quebec, wuli@us.ibm.com}\\
{\tt mrinmaya.sachan@inf.ethz.ch, wlh@cs.mcgill.ca}
}

\date{}

\begin{document}

\maketitle


\begin{abstract}
Prior work on automated question generation has almost exclusively focused on generating simple questions whose answers can be extracted from a single document.
However, there is an increasing interest in developing systems that are capable of more complex multi-hop question generation, where answering the questions requires reasoning over multiple documents.
In this work, we introduce a series of strong transformer models for multi-hop question generation, including a graph-augmented transformer that leverages relations between entities in the text. 
While prior work has emphasized the importance of graph-based models, we show that we can substantially outperform the state-of-the-art by {5 BLEU points}  using a standard transformer architecture. We further demonstrate that graph-based augmentations can provide complimentary improvements on top of this foundation.
Interestingly, we find that several important factors---such as the inclusion of an auxiliary contrastive objective and data filtering could have larger impacts on performance. 
We hope that our stronger baselines and analysis provide a constructive foundation for future work in this area.
\end{abstract}


\section{Introduction}
Motivated by the process of human inquiry and learning, the field of question generation (QG) requires a model to generate natural language questions in context.
QG has wide applicability in automated dialog systems~\cite{mostafazadeh2016generating,woebot2017}, language assessment~\cite{settles2020asessment}, data augmentation~\cite{tang2017question}, and the development of annotated data sets for question answering (QA) research.

\begin{figure}[t!]
    \centering
\small
\def\arraystretch{1.5}
\begin{tabular}{ | p{0.43\textwidth} | }
\hline
{\bf Document 1}: \textit{Byron Edmund Walker} \newline
[\textbf{1}] \dorange{Sir \underline{Byron Edmund Walker}, CVO (14 October 1848 – 27 March 1924) was a Canadian banker.} [\textbf{2}] \dorange{He was the president of the Canadian Bank of Commerce from 1907 to 1924, and a generous patron of the arts, helping to \underline{found} and nurture many of Canada's cultural and educational \underline{institutions}, including the \underline{University of Toronto}, National Gallery of Canada, \ldots} 
\vspace{0.1pt}
\newline
{\bf Document 2}: \textit{University of Toronto} \newline
[\textbf{1}] \dblue{\underline{The University of Toronto} (U of T, UToronto, or Toronto) is a public research university in Toronto, \ldots} 
[2] It was founded by royal charter in 1827 as ``King's College'', the first institution of higher learning \ldots 
[\textbf{3}] \dblue{\underline{Originally controlled} by the \underline{Church of England}, the university assumed the present name in 1850 upon becoming a secular institution.} [4] As a \ldots 
\\
\hline
{\bf Answer}: The University of Toronto
\\ 
\hline
{\bf Supporting Facts} \newline
Document 1: \{1, 2\}, Document 2: \{1, 3\}
\\ 
\hline
{\bf Question} \newline
Which \textit{Byron Edmund Walker founded institution} was \textit{originally controlled} by the \textit{Church of England}?
\\
\hline

\end{tabular}

    \caption{A (truncated) example illustrating the multi-hop QG task. The inputs are the two documents, answer, and supporting facts. The model is expected to generate a multi-hop question such that it is answerable using both the documents together. Entities and predicates relevant to generating the question are underlined in the documents, supporting facts are shown in color and their sentences ids are highlighted in bold.}
    \label{fig:intro-example}
    \vspace{-15pt}
\end{figure}

Most prior research on QG has focused on generating relatively simple {\em factoid-based} questions, where answering the question simply requires extracting a span of text from a single reference document~\cite{zhao2018paragraph,kumar2019question}.
However, motivated by the desire to build NLP systems that are capable of more sophisticated forms of reasoning and understanding~\cite{kaushik2018much, sinha2019clutrr}, there is an increasing interest in developing systems for {\em multi-hop} question answering and generation \cite{zhang2017variational, welbl2018constructing, yang2018hotpotqa, dhingra2020differentiable}, where answering the questions requires reasoning over the content in multiple text documents (see Figure~\ref{fig:intro-example} for an example).

Unlike standard QG, generating multi-hop questions requires the model to understand the relationship between disjoint pieces of information in multiple context documents. 
Compared to standard QG, multi-hop questions tend to be substantially longer, contain a higher density of named entities, and---perhaps most importantly---high-quality multi-hop questions involve complex chains of predicates connecting the mentioned entities (see Appendix \S\ref{sec:ques_complexity} for supporting statistics.)

To address these challenges, existing research on multi-hop QG primarily relies on graph-to-sequence (G2S) methods~\cite{pan2020semantic,Yu:20}.
These approaches extract graph inputs by augmenting the original text with structural information (e.g., entity annotations and dependency parses) and then apply graph neural networks (GNNs)~\cite{kipf2016semi,hamilton2017representation} to learn graph embeddings that are then fed to a sequence-based decoder. 
However, the necessity of these complex G2S approaches---which require designing hand-crafted graph extractors---is not entirely clear, especially when standard transformer-based sequence-to-sequence (S2S) models already induce a strong relational inductive bias~\cite{vaswani2017attention}.
Since transformers have the inherent ability to reason about the relationships between the entities in the text, one might imagine that these models alone would suffice for the relational reasoning requirements of multi-hop QG. 

\xhdr{Present work}
In this work, we show that, in fact, a standard transformer architecture is sufficient to outperform the prior state-of-the-art on multi-hop QG. 
We also propose and analyze a graph-augmented transformer (GATE)---which integrates explicit graph structure information into the transformer model.
GATE sets a new state-of-the-art and outperforms the best previous method by 5 BLEU points on the HotpotQA dataset~\cite{yang2018hotpotqa}.
However, we show that the gains induced by the graph augmentations are relatively small compared to other improvements in our vanilla transformer architecture, such as an auxiliary contrastive objective and a data filtering approach, which improve our model by 7.9 BLEU points in ablation studies. 
Overall, our results suggest diminishing returns from incorporating hand-crafted graph structures for multi-hop reasoning and provides a foundation for stronger multi-hop reasoning systems based on transformer architectures. 

\noindent
Our key contributions are summarized as follows:
\begin{itemize}[leftmargin=*,itemsep=1pt,topsep=2pt,parsep=2pt]
    \item We propose a strong transformer-based approach for multi-hop QG, achieving new state-of-the-art performance without leveraging hand-crafted graph structures. 
    \item We further show how graph augmentations can be integrated into the transformer architecture, leading to an overall increase of 5 BLEU points compared to previously published work. 
    \item Detailed ablations and error analysis highlight essential challenges of multi-hop QG---such as distributional mismatches---that have largely gone unnoticed in previous work and reveal critical design decisions (e.g., for data filtering). 
\end{itemize}
We hope that our work provides a strong foundation for future research on multi-hop QG while guiding the field towards the most promising avenues for future model improvements.


\section{Methods} \label{sec:methods}

In this section, we formalize the multi-hop question generation (QG) task and introduce a series of strong transformer-based models for this task. 
In particular, we first describe how we adapt the standard transformer architecture proposed by \citet{vaswani2017attention} to multi-hop QG (\S\ref{sub:te}).
Following this, we introduce an approach for augmenting a transformer with graph-structured information (\S\ref{sub:gate}), and, finally, we outline two techniques that are critical to achieving strong performance: an auxiliary contrastive objective (\S\ref{sec:composite_obj}) and a data filtering approach (\S\ref{sub:qlen_dist}). 

\paragraph{Problem Formulation} 
The input to the multi-hop QG task is a set of context documents $\{c_1,..,c_k\}$ and an answer $a$. These documents can be long containing multiple sentences, i.e., $c_j = [s_1,...,s_n]$, where each $s_i = [w^{(i)}_1, \dots, w^{(i)}_t]$ is composed of a sequence of tokens. 
Sentences across different documents are linked through bridge entities, which are named entities that occur in multiple documents.
The answer $a$ always spans one or multiple tokens in one document. 
The desired goal of multi-hop QG is to generate a question $q$ conditioned on the context and the answer, where answering this question requires reasoning about the content in more than one of the context documents. 

\subsection{Sequence-to-Sequence via Transformers}\label{sub:te}

Our base architecture for multi-hop QG is a transformer-based sequence-to-sequence (S2S) model \cite{vaswani2017attention}. 
In particular, we formulate multi-hop QG as a S2S learning task, where the input sequence contains the concatenation of the context documents $[c_1,..,c_k]$ and the provided answer $a$. 
In a transformer S2S model, both the encoder and decoder consist of self-attention and feed-forward sublayers, which are trained using teaching-forcing and a negative log-likelihood loss \cite{williams1989tf}. 
We describe the basic self-attention and feed-forward sublayers below.
In addition, we found that achieving strong performance with a transformer required careful design decisions in terms of how the input is annotated in the encoder and decoder, so we include a detailed description of our input annotation technique.

\subsubsection{Transformer sublayers}
\paragraph{Self-attention sublayer}
The self-attention sublayer performs \emph{dot-product self-attention}. Let the input to the sublayer be token embeddings $x = (x_1,\ldots,x_{\textit{T}})$ and the output be $z = (z_1,\ldots,z_{\textit{T}})$, where $x_i, z_i \in \mathbb{R}^{d}$. First, the input is linearly transformed to obtain key ($k_i=x_i \boldsymbol{W_{\textit{K}}}$), value ($v_i=x_i \boldsymbol{W_{\textit{V}}}$), and query ($q_i=x_i \boldsymbol{W_{\textit{Q}}}$) vectors. Next, interaction scores ($s_{ij}$) between query and key vectors are computed by performing a dot-product operation $s_{ij} = q_i k_j^{\top}$.
Then, attention coefficients ($\alpha_{ij}$) are computed by applying softmax function over these interaction scores $\alpha_{ij} = \frac{\exp{s_{ij}}}{\sum_{l=1}^{\textit{T}}\exp{s_{il}}}.$
Finally, self-attention embeddings ($z_i$) are computed by the weighted combination of attention coefficients with value vectors followed by a linear transformation $z_i = (\sum_{j=1}^T\alpha_{ij}v_j) \boldsymbol{W_{\textit{F}}}$.

\paragraph{Feed-forward sublayer}
In feed-foward sublayer, we pass as input the embeddings of all the tokens to a two-layer $\mathrm{MLP}$ with $\mathrm{ReLU}$ activation. $h_i = \max(0,\:z_i \boldsymbol{W_{\textit{L}_1}} + b_1) \boldsymbol{W_{\textit{L}_2}} + b_2$,
where $\boldsymbol{W_{\textit{L}_1}}\in\mathbb{R}^{d \times d'}$, $\boldsymbol{W_{\textit{L}_2}}\in\mathbb{R}^{d' \times d}$. These embeddings ($h_i$) are given as input to the next layer.

In the above descriptions, all the weight matrices (denoted by $\boldsymbol{W_*}$) and biases (denoted by $b_*$) are trainable parameters. 

\subsubsection{Input annotations}
\paragraph{Sentence and document annotations}
As the sentences present in the document context are expected to play a crucial role in learning, we add additional annotations to the input in order to learn sentence-level embeddings.
Learning these sentence-level embeddings adds a form of implicit regularization, and we also leverage these embeddings in our auxiliary contrastive loss (\S\ref{sec:composite_obj}). 
In particular, we add a \emph{sentence id token} after the last token of a sentence.
We similarly use special tokens to represent each document.
In practice, as the number of sentences varies between examples we \emph{tie} the sentence id token embedding weights for all sentences and refer to it as the \textit{<sep>} token. This simple trick also makes the model more robust to the training-test distribution shift arising due to the difference in the number of sentences in context.

\paragraph{Annotating the answer span}
To provide the answer tokens as input to the encoder, the prevalent technique in QG approaches is to append the answer tokens after the context tokens with a delimiter token between them~\cite{dong2019unified}. However, we found this approach to substantially under-perform in multi-hop QG. A possible reason is that answer tokens concatenation imparts poor inductive biases to the decoder. To overcome this limitation, we define indicator \emph{answer type id} tokens in which the value of type ids is \texttt{1} for the answer span tokens (within the context) and \texttt{0} for the remaining tokens. We introduce a new embedding layer for the answer type ids, and the \emph{answer type embeddings} are added to the context token embeddings. 

\paragraph{Delimiters in the decoder}
Finally, for the decoder input, in addition to the question tokens, we define two special tokens: \textit{<bos>} and \textit{<eos>}. This is done to simplify the question generation step during the decoding process. To decode the question sequence during inference, we initially feed the \textit{<bos>} token to the decoder and stop the generation process when the \textit{<eos>} token is emitted.

\subsection{Graph-Augmented Transformer Encoder}\label{sub:gate}

Having described our basic transformer approach, we now discuss how this transformer can be augmented by extracting explicit graph-structure from the input context. 
In addition to the document-level structure such as paragraphs and sentences, the context also contains structural information such as \emph{entities} and \emph{relations} among them, and a popular approach in the multi-hop setting is to use graph neural networks (GNNs) to encode this structural information \cite{pan2020semantic}.
In this work, we augment the transformer architecture itself with the graph-structure information---an approach that we found to substantially outperform other graph-to-sequence alternatives. 
We refer to this approach as the graph-augmented transformer encoder (GATE) and contrast it with the basic transformer encoder (TE) discussed in the previous section. 

\subsubsection{Graph Representation of Documents}

\begin{figure}
    \centering
    \includegraphics[width=0.4\textwidth]{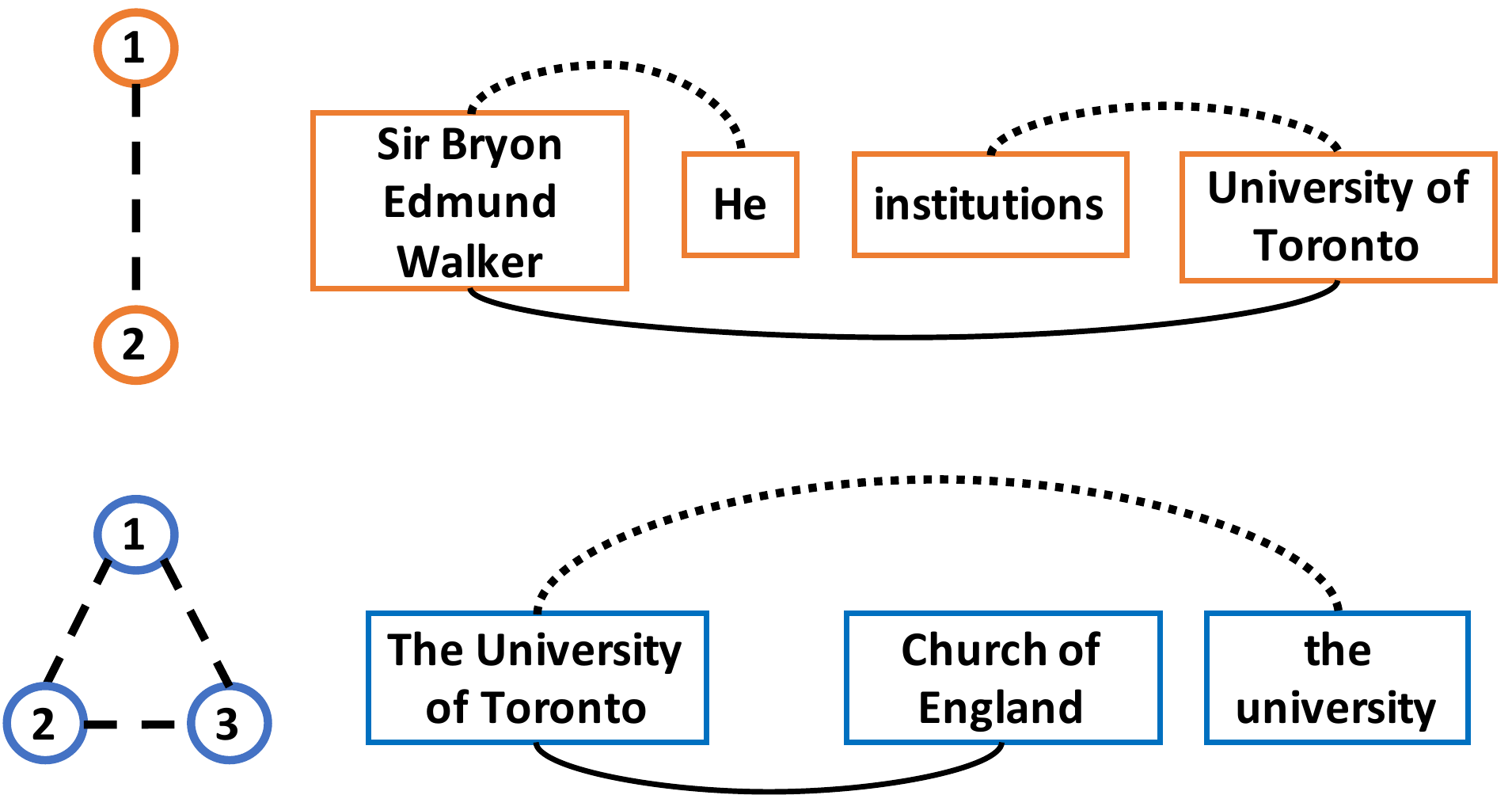}
    \caption{Entity-centric graph corresponding to the example in Figure~\ref{fig:intro-example}. The sentence nodes are drawn in circles and the entities in rectangles. Entity edges are drawn in bold, coreference edges are dotted and sentence edges are dashed.}
    \label{fig:graph-example}
\end{figure}

To extract graph structure from the input context, we consider three types of nodes---\emph{named-entity mentions}, \emph{coreferent-entities}, and \emph{sentence-ids}---and we extract a \emph{multi-relational graph} with three types of relations over these nodes (Figure \ref{fig:graph-example}). First, we extract named entities present in the context and introduce edges between them.\footnote{We use the English NER model provided by Spacy toolkit, which was trained on OntoNotes-5.0 and covers 18 classes.} Second, we extract coreferent words in a document and connect them with edges.\footnote{We use the coreference resolution model trained on OntoNotes-5.0 provided by Spacy.} Third, we introduce edges between all sentence nodes in the context.
As entities comprise the nodes of this graph, we refer to it as ``\emph{context-entity graph}''.

\subsubsection{GATE Sublayers}

We leverage the context-entity graph by defining two new types of transformer sublayers: a {\em graph-attention sublayer} and a {\em fused-attention sublayer}.
These two sublayers are intended to be used in sequence with each other and in conjunction with the usual self-attention and fully-connected sublayers of a transformer.

\paragraph{Graph-attention sublayer}
The graph-attention sublayer performs \emph{relational dot-product graph-attention}. The input to this sublayer are node embeddings\footnote{The node embeddings are obtained from the entity's token embeddings.} from the context-entity graph $v = (v_1,\ldots,v_{\textit{N}})$. Here, we aggregate information from the connected nodes instead of all the tokens. First, interaction scores ($\tilde{s}_{ij}$) are computed for all the edges by performing dot-product on the adjacent nodes embeddings $\tilde{s}_{ij} = (v_i \boldsymbol{\widetilde{W}_{\textit{Q}}}) (\tilde{v}_j \boldsymbol{\widetilde{W}_{\textit{K}}} + \gamma_{ij})^{\top}$.
In this step, we additionally account for the relation between the two nodes by learning embeddings $(\gamma \in \mathbb{R}^d)$ for each relation type~\cite{shaw2018self}, where $\gamma_{ij}$ denotes the relation type between nodes $i$ and $j$. Next, we compute attention score ($\tilde{\alpha}_{ij}$) for each node by applying softmax over the interaction scores from all its connecting edges $\tilde{\alpha}_{ij} = \frac{\exp(\tilde{s}_{ij})}{{\sum_{k\in \mathcal{N}_i}\exp{(\tilde{s}_{ik})}}}$,
where $\mathcal{N}_\textit{i}$ refers to the set of nodes connected to the $i^{\textit{th}}$ node. Graph-attention embeddings ($\tilde{z}_i$) are computed by the aggregation of attention scores followed by a linear transformation $\tilde{z}_i = (\sum_{j\in \mathcal{N}_i}\tilde{\alpha}_{ij}(\tilde{v}_j \boldsymbol{\widetilde{W}_{\textit{V}}} + \gamma_{ij}))\boldsymbol{\widetilde{W}_{\textit{F}}}$.

\paragraph{Fused-attention sublayer}
After running both the graph-attention sublayer described above, as well as the standard self-attention sublayer described in \S\ref{sub:te}, the context tokens which belong to the vertex set of context-entity graph will have two embeddings: $z_i$ from self-attention and $\tilde{z}_{i}$ from graph-attention. To effectively integrate information from sequence- and graph-views, we concatenate these two embeddings and apply a parametric function $f$ such as $\mathrm{MLP}$ with $\mathrm{ReLU}$ non-linearity~\cite{glorot2011deep}, which we term as the \emph{fused-attention sublayer} $z_i = f(\:[z_i,\:\tilde{z}_i])$, where $z_i \in \mathbb{R}^d$.

\subsection{Training Losses} \label{sec:composite_obj}
To train our S2S transformer models, we combine two loss functions. 
The first loss function is the standard S2S log-likelihood loss, while the second loss function trains the model to detect useful sentences within the question context. 

\subsubsection{Negative Log-Likelihood Objective}
Our primary training signal for our sequence-to-sequence approach comes from a standard negative log-likelihood loss:
\begin{align*}
\sL_{\text{NLL}} = -\frac{1}{K} \sum_{k=1}^{K} \log{p\left(q_{k} \mid c, q_{1:k-1}\right)},
\end{align*}
where the parametric distribution $p(q \mid c)$ models the conditional probability of question ($q$) given the context ($c$) and $K$ is the number of question tokens.
As is common practice in the literature, we use teacher-forcing while training with this loss. 

\subsubsection{Auxiliary Contrastive Objective}

To compliment our standard likelihood objective, we also design a contrastive objective, which trains the model to detect the occurrence of {\em supporting facts} in the multi-document context. 

\paragraph{Supporting facts in multi-hop QA}
One of the unique challenges of multi-hop QA is the fact that the context contains a large number of irrelevant sentences due to the fact that multi-hop questions require reasoning over multiple long documents.
Thus, as a common practice, researchers will annotate which sentences are necessary to answer each question, called {\em supporting facts} \cite{yang2018hotpotqa}.
Prior work on multi-hop QG leveraged these annotations by simply discarding all irrelevant sentences and training only on sentences with supporting facts \cite{yang2018hotpotqa}.
Instead, we propose a contrastive objective, which allows us to leverage these annotations during training while still receiving full document contexts at inference. 

\paragraph{Contrastive objective}
Our contrastive learning setup utilizes sentences contained in the supporting facts as positive examples ($y=1$) while we consider all the remaining sentences in the context to be negative examples ($y=0$). We  use only the sentence id embedding for training (\emph{i.e.}, the embedding corresponding to the \emph{<sep>} token; see \S\ref{sub:te}) but not the words contained in the sentence. Let $h_i$ denote the sentence id embedding.
The contrastive training loss is defined in terms of binary cross-entropy loss formulation as:
\begin{align*}
\sL_{\text{CT}} &= -\frac{1}{P+N}\left(\sum_{i=1}^P \mathbbm{1}(y_i=1)\log \sD\left(h_i \right)\right. \\
&+ \left. \sum_{j=1}^N \mathbbm{1}(y_i=0)\log\left(1 - \sD\left( {h}_j \right)\right)\right),
\end{align*}
where $\mathcal{D}$ is a binary classifier consisting of a two-layer MLP with ReLU activation and a final sigmoid layer, $P$ and $N$ are the number of positive and negative training sentences in the context documents respectively.
During evaluation, we predict the supporting facts using the binary classifier and also calculate the F1 and Exact Match (EM) scores from the predictions.
This contrastive objective is added as a regularization term in addition to the main likelihood loss, leading to the following composite objective: 
\begin{align*}
\sL = \lambda \sL_{\text{CT}} + (1-\lambda) \sL_{\text{NLL}}.
\end{align*}

\subsection{Data Filtering Approach}\label{sub:qlen_dist}

\begin{figure}[t]
\centering
\makebox[\linewidth][c]{\includegraphics[max width=1\linewidth, scale=1.0]{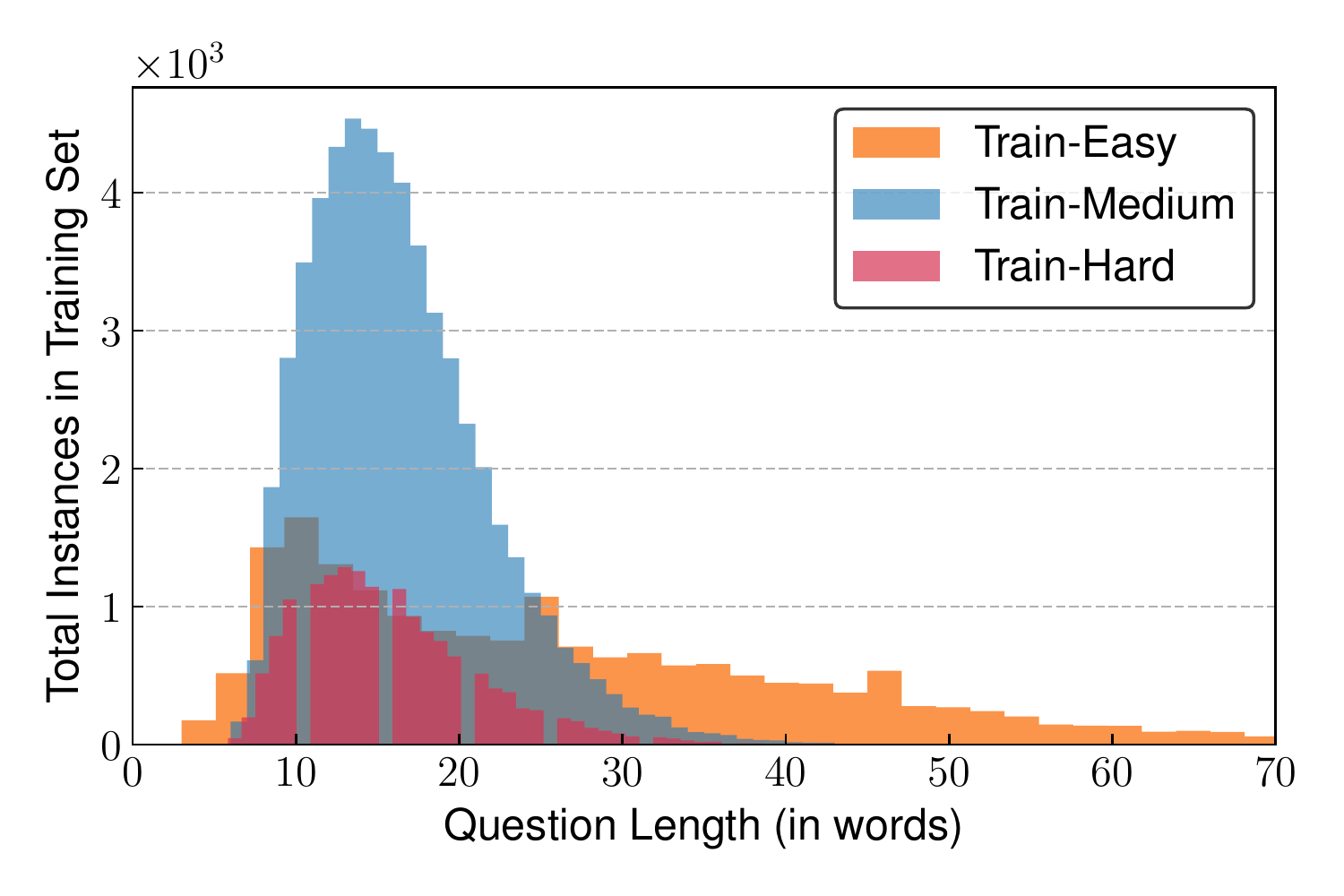}}
\caption{Question length distribution according to its difficulty level in the HotpotQA training set. Plot reveals that train-easy questions are much longer than train-medium and train-hard questions.}
\label{fig:ques_length}
\end{figure}

The final key component of our transformer-based multi-hop QG model is a data filtering approach.
This aspect of our model specifically addresses challenges arising from the question-length distribution in the standard HotpotQA benchmark \cite{yang2018hotpotqa}, which is the main multi-hop QA dataset analyzed in this work. 
Nonetheless, despite the fact that this approach is motivated directly by the statistics of HotpotQA, we expect the general principle to be applicable to future multi-hop datasets as well. 

\paragraph{The question-length distribution in HotPotQA}
The training set of HotpotQA consists of three categories: \emph{train-easy}, \emph{train-medium}, and \emph{train-hard}. Train-easy questions are essentially single-hop; i.e., they need one context document to extract the answer while both train-medium and train-hard questions are multi-hop requiring multiple context documents. \emph{However, both the dev and test sets in HotpotQA mainly consist of hard multi-hop questions}. 
While the additional {\em train-easy} and {\em train-medium} examples have proved useful as training signals in the question-answering setting, our QG experiments reveal that naively using the provided training distribution of questions leads to a significant drop in BLEU scores on the development set. The reason for the lower BLEU scores is that the generated questions are almost 80\% longer than the reference questions, and thus are less precise. 

\paragraph{Filtering to avoid distributional mismatch}
We speculate that the model generates long questions because of the \emph{negative exposure bias} which it receives due to train-easy questions being much longer than train-medium and train-hard. We plot the distribution of the question length in the training set in Figure~\ref{fig:ques_length}. We observe that a significant number of train-easy questions are much longer than train-medium and train-hard---while most of the train-medium and train-hard questions are 30 words long, train-easy questions can be as long as 70 words.
Thus, we match the training-dev question-length distribution by pruning examples whose question length is more than 30 words in our training set. According to our analysis above, most of these pruned questions are train-easy questions.
Although one can adopt complex data-weighting techniques for this~\cite{zhiting2019nips}, we observed that simple hard-filtering works well in practice in our case.


\section{Experimental Setup} \label{sec:setup}

\subsection{Dataset Preprocessing and Evaluation} \label{sub:preprocessing}
We use HotpotQA dataset~\cite{yang2018hotpotqa} for experiments as it is the only multi-hop QA dataset that contains questions in textual form.\footnote{We also explored WikiHop~\cite{welbl2018constructing} but it contains questions in triple format and thus is outside the scope of this work.} HotpotQA is a large-scale crowd-sourced dataset constructed from Wikipedia articles and contains over 100K questions. We use its \emph{distractor} setting that contains 2 gold and 8 distractor paragraphs for a question. Following prior work on multi-hop QG, we limit the context size to the 2 gold paragraphs, as the distractor paragraphs are irrelevant to the generation task \cite{pan2020semantic}. 
The questions can be either of type \emph{bridge}- or \emph{comparison}-based. The answer span is not explicitly specified in the context documents rather the answer tokens are provided. Hence, we use approximate text-similarity algorithms to search for the best matching answer span in the context. For some of the comparison questions whose answer is either \emph{yes} or \emph{no}, we append it to the context.

To train and evaluate the models, we use the standard training and dev sets.\footnote{As the test set is hidden for HotpotQA.} We pre-process the dataset by excluding examples with spurious annotations and filter out training instances whose question length is more than 30 words. As the official dev set is used as a test set, we reserve 500 examples from the training set to be used as dev set. Overall, our training set consists of 84,000 examples, and the test set consists of 7,399 examples.
We follow the evaluation protocol of \citet{pan2020semantic} and report scores on standard automated evaluation metrics common in QG: BLEU~\cite{papineni2002bleu},\footnote{This is also known as BLEU-4.} ROUGE-L~\cite{lin2004rouge}, and METEOR~\cite{banerjee2005meteor}.

\subsection{Training Protocols}
For all the experiments, we follow the same training process. We encode the context and question words with subwords units by applying a \emph{unigram language model} as implemented in the open-source \textit{sentencepiece} toolkit~\cite{kudo2018sentencepiece}. We use 32,000 subword units including 4 special tokens (\textit{<bos>}, \textit{<eos>}, \textit{<sep>}, \textit{<unk>}).
The first three of these subword units were introduced in \S\ref{sec:methods}, while the \textit{<unk>} or \textit{unknown} token helps to scale to larger vocabularies and provides a mechanism to handle new tokens at test time. 

For all the experiments, we use a 2-layer transformer model with 8 attention heads, 512-D model size, and 2048-D hidden layer.\footnote{This is the Transformer-\emph{base} setting from the original paper, apart from the number of layers.}  
Word embedding weights are shared between the encoder, decoder, and generation layer. For reproducibility, we describe model training details in Appendix~\ref{sec:train_det}.


\section{Results and Analysis}  \label{sec:results}

\begin{table}[t]
\centering
\begin{tabular}{@{}l r r r@{}}
 \toprule
 \textbf{Model} & \textbf{BLEU}  & \textbf{ROUGE-L} & \textbf{METEOR} \\
 \midrule
 \multicolumn{4}{c}{\textit{Encoder Input: Supporting Facts Sentences}}\\
 \midrule
 NQG++$^\dagger$ & 11.50 & 32.01 & 16.96 \\
 ASs2s$^\dagger$ & 11.29 & 32.88 & 16.78 \\
 MP-GSA$^\dagger$ & 13.48 & 34.51 & 18.39 \\
 SRL-Graph$^\dagger$ & 15.03 & 36.24 & 19.73 \\
 DP-Graph$^\dagger$ & 15.53 & 36.94 & 20.15 \\
 GATE$_{\text{NLL}}$ & \bf{19.33} & \bf{39.00} & \bf{22.21} \\
 \midrule
 \midrule
 \multicolumn{4}{c}{\emph{Encoder Input: Full Document Context}}\\
 \midrule
 TE$_{\text{NLL+CT}}$ & \bf{19.60} & \bf{39.23} & \bf{22.50}\\
 GATE$_{\text{NLL}}$ & 17.13 & 38.13 & 21.34 \\
 GATE$_{\text{NLL+CT}}$ & \bf{20.02} & \bf{39.49} & \bf{22.40} \\
 \bottomrule
 \end{tabular}
\caption{Results of multi-hop QG on HotpotQA. NQG++ is from~\citet{zhou2018nqg}, ASs2s is from~\citet{kim2019improving}, MP-GSA is from~\citet{zhao-etal-2018-paragraph}, SRL-Graph and DP-Graph are from~\citet{pan2020semantic}. $\dagger$ denotes that the results are taken from~\citet{pan2020semantic}. Best results in each section are highlighted in bold.}
\label{tab:model_bleu}
\end{table}

We report the performance of our proposed transformer encoder (TE) and graph-augmented transformer encoder (GATE) models in Table~\ref{tab:model_bleu}, comparing with a number of recent QG approaches. In subsequent discussions, we will mostly use the BLEU values from this table to compare performance, but we also provide ROUGE-L and METEOR scores.

\paragraph{Performance with supporting facts as input}
We first consider a \emph{simplified version} of the task when \emph{only the supporting facts are used during training and testing} (top section in Table~\ref{tab:model_bleu}).
In other words, in this setting, we remove all sentences of the context documents that have not been annotated as supporting facts.
This is an overly simplified setting since supporting fact annotations are not always available at test time.
However, this is the setting used in previous work on multi-hop QG \cite{pan2020semantic}, which we directly compare to. 
In this setting, we see that our GATE models scores 19.33 BLEU, an absolute gain of around 4 points over the previous best result.
Note that as contrastive training is not applicable here, we just use the log-likelihood loss ($\sL_{\text{NLL}}$) for training. 

\begin{figure}[t]
\centering
\makebox[\linewidth][c]{\includegraphics[max width=1\linewidth, scale=1.0]{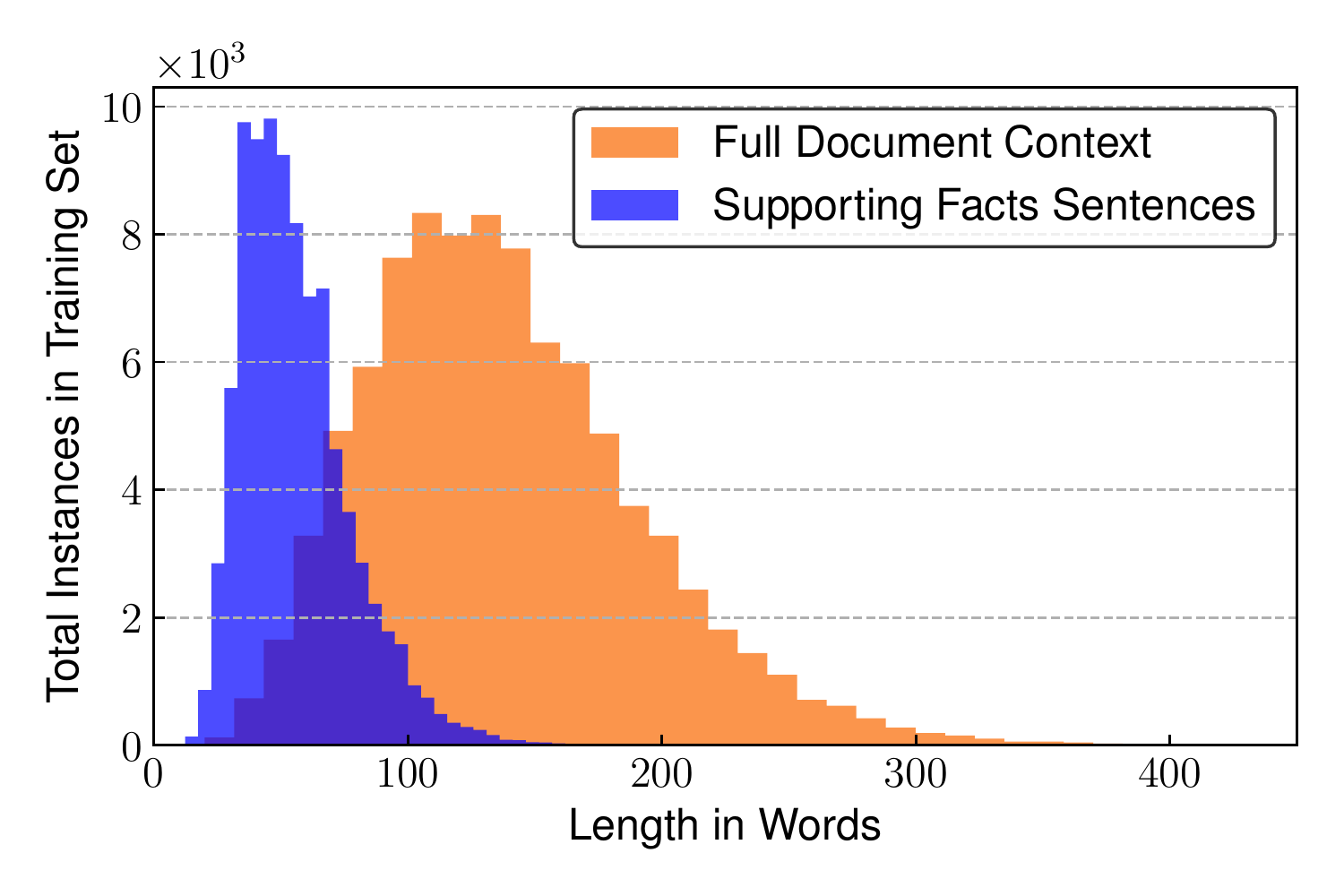}}
\caption{Length distribution of full document context and supporting facts sentences in HotpotQA. It reveals that the full document context is almost three times longer than supporting facts.}
\label{fig:cl}
\end{figure}

\paragraph{Performance with full context as input}
In a more realistic setting when the supporting facts are not available at test time, the model needs to processes the full context. As the average document context is three times the size of the supporting facts in HotpotQA (Figure~\ref{fig:cl}), this setting is potentially much more challenging. As is evident from results (bottom section in Table~\ref{tab:model_bleu}), we see that the BLEU of our GATE model using only the likelihood objective GATE$_{\text{NLL}}$ drops to 17.13, an absolute drop of nearly 2 points. However, when trained using the composite objective (GATE$_{\text{NLL+CL}}$), the GATE model is able to obtain a BLEU of 20.02, which is actually higher than what GATE could achieve in the simplified setting.
We suspect that the additional training signal from the longer contexts combined with the contrastive objective actually benefits the models. 
Moreover, we find that our TE model---which contains no graph augmentations---is also able to achieve very strong performance in this setting, achieving a BLEU of 19.60, which also substantially outperforms all previous methods. 

\subsection{Ablation Studies} \label{sub:ans_rep}

\begin{table}[t]
\centering
\small
\begin{tabular}{l r@{}}
 \toprule
 \textbf{Setting} & \textbf{BLEU} \\
 \midrule
 GATE$_{\text{NLL+CT}}$ & 20.02 \\
 \midrule
 -- \emph{contrastive training} & 17.13 \\
 \midrule
 -- \emph{data filtering} & 14.50 \\
 -- \emph{data filtering, contrastive training} & 11.90 \\
 \midrule
 -- \emph{answer type ids} & 7.81 \\
 \bottomrule
 \end{tabular}
\caption{Ablation studies when the encoders' input is the full document context.}
\label{tab:input_rep}
\end{table}

We perform ablation studies to understand what components are essential for strong performance on multi-hop QG (Table~\ref{tab:input_rep}).

\paragraph{Contrastive Training}
We see that contrastive training improves the BLEU score by 3 points and it also helps the model attain around 75 F1 and 35 Exact Match scores in predicting the supporting facts sentences. This highlights that---besides being helpful in QG---contrastive training additionally imparts useful signals to the encoder such that its supporting facts predictions are \emph{interpretable}.

\paragraph{Data Filtering}
Next, we see that data filtering of easy questions greater than 30 words is a critical step in our training pipeline.
If we use the full training data, the accuracy drops by about 5 points to 14.5 BLEU. We also observed the generations to be 1.6-1.8x longer than the reference questions. Furthermore, if we just use teacher-forcing on the entire training set, then the performance further drops to 11.90 BLEU, signifying that question filtering and contrastive training are independently useful and essential. 

\paragraph{Answer Encoding}
Finally, we show that effective encoding of the answer is of utmost importance in multi-hop QG as the decoder needs to condition generation on both the context and answer. As mentioned previously, the common approach in standard QG is to append the answer tokens after the context. We see that using this approach results in a drop of around 12 points to 7.81 BLEU, which is quite low.
Our approach to marking the answer span in the context with answer type ids appears to be a much stronger methodology. 

\subsection{Complementarity of TE and GATE} \label{sub:ensemble}

\begin{table}[t]
\centering
\begin{tabular}{@{}l r r r@{}}
 \toprule
 \textbf{Model} & \textbf{BLEU}  & \textbf{ROUGE-L} & \textbf{METEOR} \\
 \midrule
 \multicolumn{4}{c}{\emph{Encoder Input: Full Document Context}}\\
 \midrule
 TE$_{\text{NLL+CT}}$ & 19.60 & 39.23 & 22.50 \\
 GATE$_{\text{NLL+CT}}$ & 20.02 & 39.49 & 22.40 \\
 Ensemble & \bf{21.34} & \bf{40.36} & \bf{23.24} \\
 \bottomrule
 \end{tabular}
\caption{Performance comparison of the TE and GATE models and their ensemble.}
\label{tab:compl_str}
\end{table}

\paragraph{Model Ensemble}
We notice that our graph-augmented model GATE seems to provide complementary strengths compared to the TE model, which is evident when we ensemble both the models during decoding. At every step, we compute the probability of the model combination using their linearly weighted probability scores as,
\begin{align*}
p\left(q_{k} \mid c, q_{1:k-1}\right) &= \alpha \cdot p_{\text{TE}}\left(q_{k} \mid c, q_{1:k-1}\right) \\
&+ (1-\alpha) \cdot p_{\text{GATE}}\left(q_{k} \mid c, q_{1:k-1}\right),
\end{align*}
where $\alpha \in [0, 1]$ is a hyperparameter.\footnote{We find that $\alpha=0.5$ works the best.} We see that the ensemble of TE and GATE model results in an accuracy score of 21.34 BLEU, which is an improvement of 1.7 points over the TE model. Ensembling also provides close to 1 point gain in the ROUGE-L scores.
This suggests that---while the gains from graph-augmentations are relatively small---there is complementary information in the explicit graph structures. 

\paragraph{GLEU Score Comparison}
In order to further understand how the GATE model is different in performance from the TE model, we perform an analysis of the generated questions on the test set. We analyze the distribution of the difference in their question-level GLEU scores~\cite{wu2016google}\footnote{GLEU is a sentence-level metric and is calculated as the minimum of the precision or recall between the reference and the hypothesis.} and observe that on 397 (5.4\%) test examples, the GATE model achieves a GLEU score of 20 points or more than that of the TE, while on 377 (5.1\%) examples the TE model achieves at least 20 points higher. Therefore, this complementary performance is the reason for the gain that we see in Table~\ref{tab:compl_str} when the two models are ensembled.


\section{Related Work} \label{rw}

Recent work on QG has mostly focused on generating one-hop questions conditioned using neural S2S models~\cite{du2017learning}, pre-trained transformers~\cite{dong2019unified}, query reformulation using reinforcement learning~\cite{buck18}, and reinforcement learning based G2S model ~\cite{chen2019reinforcement}.
Contemporary to our work,~\cite{pan2020semantic,Yu:20} also propose approaches for multi-hop QG. Similar to our work, these works incorporate an entity-graph to capture information about entities and their contextual relations within as well as across multiple documents. In addition to modeling the entity-graph, our approach also uses contrastive training with teacher-forcing to allow the model to efficiently use the information presented in the supporting facts.

In parallel, there have been advances in multi-hop question answering (as opposed to generation) models \cite{tu2019select,chen2019multihop,tu2020graph,groeneveld2020simple}. GNN models applied over the extracted graph structures have led to improvements in this domain ~\cite{de2019question, fang2019hierarchical,zhao2020transformer-xh}. 
Our work examines the complementary task of multi-hop QG and provides evidence that stronger transformer models could, in fact, achieve more competitive results in this domain, compared to these GNN-based models that use explicit graph structure. 

Also related to our work is the recent line of work on graph-to-text transduction \cite{xu2018graph2seq,koncel2019text,zhu2019modeling,cai2019graph,chen2019reinforcement}. 
However, these works seek to generate text from a structured input, rather than the setting we examine, which involves taking context text as the input.


\section{Conclusion} \label{sec:conclusion}
In this work, we propose a series of strong transformer models for multi-hop QG. To effectively encode the context documents and the answer, we introduce answer type embeddings and a new sublayer to incorporate the extracted entity-centric graph. We also propose an auxiliary contrastive objective to identify the supporting facts and a data filtering approach to balance the training-test distribution mismatch. Experiments on the HotpotQA dataset show that our models outperform the current best approaches by a substantial margin of 5 BLEU points. Our analysis further reveals that graph-based components may not be the most critical in improving the performance, but can render complementary strengths to the transformer.

\bibliography{multi_hop_qa_gen}

\begin{thebibliography}{51}
\expandafter\ifx\csname natexlab\endcsname\relax\def\natexlab#1{#1}\fi

\bibitem[{Banerjee and Lavie(2005)}]{banerjee2005meteor}
Satanjeev Banerjee and Alon Lavie. 2005.
\newblock {METEOR}: An automatic metric for {MT} evaluation with improved
  correlation with human judgments.
\newblock In \emph{Proceedings of the {ACL} Workshop on Intrinsic and Extrinsic
  Evaluation Measures for Machine Translation and/or Summarization}, pages
  65--72, Ann Arbor, Michigan. Association for Computational Linguistics.

\bibitem[{Buck et~al.(2018)Buck, Bulian, Ciaramita, Gesmundo, Houlsby,
  Gajewski, and Wang}]{buck18}
Christian Buck, Jannis Bulian, Massimiliano Ciaramita, Andrea Gesmundo, Neil
  Houlsby, Wojciech Gajewski, and Wei Wang. 2018.
\newblock Ask the right questions: {{Active}} question reformulation with
  reinforcement learning.
\newblock In \emph{Sixth International Conference on Learning Representations
  ({{ICLR}})}, {Vancouver, Canada}.

\bibitem[{Cai and Lam(2020)}]{cai2019graph}
Deng Cai and Wai Lam. 2020.
\newblock Graph transformer for graph-to-sequence learning.
\newblock In \emph{Proceedings of The Thirty-Fourth AAAI Conference on
  Artificial Intelligence (AAAI)}.

\bibitem[{Chen et~al.(2019)Chen, Lin, and Durrett}]{chen2019multihop}
Jifan Chen, Shih-ting Lin, and Greg Durrett. 2019.
\newblock Multi-hop question answering via reasoning chains.
\newblock \emph{arXiv preprint arXiv:1910.02610}.

\bibitem[{Chen et~al.(2020)Chen, Wu, and Zaki}]{chen2019reinforcement}
Yu~Chen, Lingfei Wu, and Mohammed~J Zaki. 2020.
\newblock Reinforcement learning based graph-to-sequence model for natural
  question generation.
\newblock In \emph{Eighth International Conference on Learning Representations
  ({{ICLR}})}.

\bibitem[{De~Cao et~al.(2019)De~Cao, Aziz, and Titov}]{de2019question}
Nicola De~Cao, Wilker Aziz, and Ivan Titov. 2019.
\newblock Question answering by reasoning across documents with graph
  convolutional networks.
\newblock In \emph{Proceedings of the 2019 Conference of the North American
  Chapter of the Association for Computational Linguistics: {{Human}} Language
  Technologies, Volume 1 (Long and Short Papers)}.

\bibitem[{Dhingra et~al.(2020)Dhingra, Zaheer, Balachandran, Neubig,
  Salakhutdinov, and Cohen}]{dhingra2020differentiable}
Bhuwan Dhingra, Manzil Zaheer, Vidhisha Balachandran, Graham Neubig, Ruslan
  Salakhutdinov, and William~W Cohen. 2020.
\newblock Differentiable reasoning over a virtual knowledge base.
\newblock In \emph{International Conference on Learning Representations}.

\bibitem[{Dong et~al.(2019)Dong, Yang, Wang, Wei, Liu, Wang, Gao, Zhou, and
  Hon}]{dong2019unified}
Li~Dong, Nan Yang, Wenhui Wang, Furu Wei, Xiaodong Liu, Yu~Wang, Jianfeng Gao,
  Ming Zhou, and Hsiao-Wuen Hon. 2019.
\newblock Unified language model pre-training for natural language
  understanding and generation.
\newblock In \emph{Advances in Neural Information Processing Systems 32}.

\bibitem[{Du et~al.(2017)Du, Shao, and Cardie}]{du2017learning}
Xinya Du, Junru Shao, and Claire Cardie. 2017.
\newblock Learning to ask: {{Neural}} question generation for reading
  comprehension.
\newblock In \emph{Proceedings of the 55th Annual Meeting of the Association
  for Computational Linguistics (Volume 1: {{Long}} Papers)}.

\bibitem[{Fang et~al.(2019)Fang, Sun, Gan, Pillai, Wang, and
  Liu}]{fang2019hierarchical}
Yuwei Fang, Siqi Sun, Zhe Gan, Rohit Pillai, Shuohang Wang, and Jingjing Liu.
  2019.
\newblock Hierarchical graph network for multi-hop question answering.
\newblock \emph{arXiv preprint arXiv:1911.03631}.

\bibitem[{Fitzpatrick et~al.(2017)Fitzpatrick, Darcy, and
  Vierhile}]{woebot2017}
Kathleen~Kara Fitzpatrick, Alison Darcy, and Molly Vierhile. 2017.
\newblock Delivering cognitive behavior therapy to young adults with symptoms
  of depression and anxiety using a fully automated conversational agent
  (woebot): A randomized controlled trial.
\newblock \emph{JMIR Ment Health}, 4(2):e19.

\bibitem[{Glorot et~al.(2011)Glorot, Bordes, and Bengio}]{glorot2011deep}
Xavier Glorot, Antoine Bordes, and Yoshua Bengio. 2011.
\newblock Deep sparse rectifier neural networks.
\newblock In \emph{Proceedings of the Fourteenth International Conference on
  Artificial Intelligence and Statistics}, volume~15 of \emph{Proceedings of
  Machine Learning Research}, {Fort Lauderdale, FL, USA}.

\bibitem[{Groeneveld et~al.(2020)Groeneveld, Khot, Sabharwal
  et~al.}]{groeneveld2020simple}
Dirk Groeneveld, Tushar Khot, Ashish Sabharwal, et~al. 2020.
\newblock A simple yet strong pipeline for hotpotqa.
\newblock \emph{arXiv preprint arXiv:2004.06753}.

\bibitem[{Hamilton et~al.(2017)Hamilton, Ying, and
  Leskovec}]{hamilton2017representation}
William~L Hamilton, Rex Ying, and Jure Leskovec. 2017.
\newblock Representation learning on graphs: {{Methods}} and applications.
\newblock \emph{arXiv preprint arXiv:1709.05584}.

\bibitem[{Hu et~al.(2019)Hu, Tan, Salakhutdinov, Mitchell, and
  Xing}]{zhiting2019nips}
Zhiting Hu, Bowen Tan, Russ~R Salakhutdinov, Tom~M Mitchell, and Eric~P Xing.
  2019.
\newblock Learning data manipulation for augmentation and weighting.
\newblock In \emph{Advances in Neural Information Processing Systems 32}.

\bibitem[{Kaushik and Lipton(2018)}]{kaushik2018much}
Divyansh Kaushik and Zachary~C. Lipton. 2018.
\newblock How much reading does reading comprehension require? {{A}} critical
  investigation of popular benchmarks.
\newblock In \emph{Proceedings of the 2018 Conference on Empirical Methods in
  Natural Language Processing}.

\bibitem[{Kim et~al.(2019)Kim, Lee, Shin, and Jung}]{kim2019improving}
Yanghoon Kim, Hwanhee Lee, Joongbo Shin, and Kyomin Jung. 2019.
\newblock Improving neural question generation using answer separation.
\newblock In \emph{Proceedings of the AAAI Conference on Artificial
  Intelligence}, volume~33, pages 6602--6609.

\bibitem[{Kingma and Ba(2014)}]{kingma2014adam}
Diederik Kingma and Jimmy Ba. 2014.
\newblock Adam: {{A}} method for stochastic optimization.
\newblock \emph{Computing Research Repository}, arXiv:1412.6980.

\bibitem[{Kipf and Welling(2016)}]{kipf2016semi}
Thomas~N Kipf and Max Welling. 2016.
\newblock Semi-supervised classification with graph convolutional networks.
\newblock \emph{arXiv preprint arXiv:1609.02907}.

\bibitem[{Koncel-Kedziorski et~al.(2019)Koncel-Kedziorski, Bekal, Luan, Lapata,
  and Hajishirzi}]{koncel2019text}
Rik Koncel-Kedziorski, Dhanush Bekal, Yi~Luan, Mirella Lapata, and Hannaneh
  Hajishirzi. 2019.
\newblock Text generation from knowledge graphs with graph transformers.
\newblock \emph{arXiv preprint arXiv:1904.02342}.

\bibitem[{Kudo and Richardson(2018)}]{kudo2018sentencepiece}
Taku Kudo and John Richardson. 2018.
\newblock {{SentencePiece}}: {{A}} simple and language independent subword
  tokenizer and detokenizer for {{Neural Text Processing}}.
\newblock In \emph{Proceedings of the 2018 Conference on Empirical Methods in
  Natural Language Processing: {{System}} Demonstrations}.

\bibitem[{Kumar et~al.(2019)Kumar, Chaki, Talluri, Ramakrishnan, Li, and
  Haffari}]{kumar2019question}
Vishwajeet Kumar, Raktim Chaki, Sai~Teja Talluri, Ganesh Ramakrishnan,
  Yuan-Fang Li, and Gholamreza Haffari. 2019.
\newblock Question generation from paragraphs: A tale of two hierarchical
  models.
\newblock \emph{arXiv preprint arXiv:1911.03407}.

\bibitem[{LeCun et~al.(1998)LeCun, Bottou, Orr, and
  M{\"u}ller}]{lecun1998efficient}
Yann LeCun, L{\'e}on Bottou, Genevieve~B Orr, and Klaus-Robert M{\"u}ller.
  1998.
\newblock Efficient backprop.
\newblock In \emph{Neural Networks: {{Tricks}} of the Trade}. {Springer}.

\bibitem[{Lin(2004)}]{lin2004rouge}
C-Y Lin. 2004.
\newblock {{ROUGE}}: {{A}} package for automatic evaluation of summaries.
\newblock In \emph{Proc. of Workshop on Text Summarization Branches out, Post
  Conference Workshop of {{ACL}} 2004}.

\bibitem[{Mostafazadeh et~al.(2016)Mostafazadeh, Misra, Devlin, Mitchell, He,
  and Vanderwende}]{mostafazadeh2016generating}
Nasrin Mostafazadeh, Ishan Misra, Jacob Devlin, Margaret Mitchell, Xiaodong He,
  and Lucy Vanderwende. 2016.
\newblock Generating natural questions about an image.
\newblock In \emph{Proceedings of the 54th Annual Meeting of the Association
  for Computational Linguistics (Volume 1: {{Long}} Papers)}.

\bibitem[{Pan et~al.(2020)Pan, Xie, Feng, Chua, and Kan}]{pan2020semantic}
Liangming Pan, Yuxi Xie, Yansong Feng, Tat-Seng Chua, and Min-Yen Kan. 2020.
\newblock Semantic graphs for generating deep questions.
\newblock In \emph{Annual Conference of the Association for Computational
  Linguistics (ACL)}.

\bibitem[{Papineni et~al.(2002)Papineni, Roukos, Ward, and
  Zhu}]{papineni2002bleu}
Kishore Papineni, Salim Roukos, Todd Ward, and Wei-Jing Zhu. 2002.
\newblock {{BLEU}}: A method for automatic evaluation of machine translation.
\newblock In \emph{Proceedings of the 40th Annual Meeting on Association for
  Computational Linguistics}. {Association for Computational Linguistics}.

\bibitem[{Pereyra et~al.(2017)Pereyra, Tucker, Chorowski, Kaiser, and
  Hinton}]{pereyra2017regularizing}
Gabriel Pereyra, George Tucker, Jan Chorowski, {\L}ukasz Kaiser, and Geoffrey
  Hinton. 2017.
\newblock Regularizing neural networks by penalizing confident output
  distributions.
\newblock \emph{Computing Research Repository}, arXiv:1701.06548.

\bibitem[{Rajpurkar et~al.(2018)Rajpurkar, Jia, and Liang}]{rajpurkar2018know}
Pranav Rajpurkar, Robin Jia, and Percy Liang. 2018.
\newblock Know what you don't know: {{Unanswerable}} questions for {{SQuAD}}.
\newblock In \emph{Proceedings of the 56th Annual Meeting of the Association
  for Computational Linguistics (Volume 2: {{Short}} Papers)}.

\bibitem[{Sachan and Neubig(2018)}]{sachan-neubig-2018-parameter}
Devendra Sachan and Graham Neubig. 2018.
\newblock Parameter sharing methods for multilingual self-attentional
  translation models.
\newblock In \emph{Proceedings of the Third Conference on Machine Translation:
  Research Papers}.

\bibitem[{Settles et~al.(2020)Settles, T.~LaFlair, and
  Hagiwara}]{settles2020asessment}
Burr Settles, Geoffrey T.~LaFlair, and Masato Hagiwara. 2020.
\newblock Machine learning–driven language assessment.
\newblock \emph{Transactions of the Association for Computational Linguistics},
  8.

\bibitem[{Shaw et~al.(2018)Shaw, Uszkoreit, and Vaswani}]{shaw2018self}
Peter Shaw, Jakob Uszkoreit, and Ashish Vaswani. 2018.
\newblock Self-attention with relative position representations.
\newblock In \emph{Proceedings of the 2018 Conference of the North {{American}}
  Chapter of the Association for Computational Linguistics: {{Human}} Language
  Technologies, Volume 2 (Short Papers)}.

\bibitem[{Sinha et~al.(2019)Sinha, Sodhani, Dong, Pineau, and
  Hamilton}]{sinha2019clutrr}
Koustuv Sinha, Shagun Sodhani, Jin Dong, Joelle Pineau, and William~L.
  Hamilton. 2019.
\newblock {CLUTRR}: A diagnostic benchmark for inductive reasoning from text.
\newblock In \emph{Proceedings of the 2019 Conference on Empirical Methods in
  Natural Language Processing and the 9th International Joint Conference on
  Natural Language Processing (EMNLP-IJCNLP)}.

\bibitem[{Srivastava et~al.(2014)Srivastava, Hinton, Krizhevsky, Sutskever, and
  Salakhutdinov}]{srivastava2014dropout}
Nitish Srivastava, Geoffrey Hinton, Alex Krizhevsky, Ilya Sutskever, and Ruslan
  Salakhutdinov. 2014.
\newblock Dropout: {{A}} simple way to prevent neural networks from
  overfitting.
\newblock \emph{Journal of Machine Learning Research}, 15(1).

\bibitem[{Stanovsky et~al.(2018)Stanovsky, Michael, Zettlemoyer, and
  Dagan}]{stanovsky-etal-2018-supervised}
Gabriel Stanovsky, Julian Michael, Luke Zettlemoyer, and Ido Dagan. 2018.
\newblock Supervised open information extraction.
\newblock In \emph{Proceedings of the 2018 Conference of the North {A}merican
  Chapter of the Association for Computational Linguistics: Human Language
  Technologies, Volume 1 (Long Papers)}.

\bibitem[{Tang et~al.(2017)Tang, Duan, Qin, Yan, and Zhou}]{tang2017question}
Duyu Tang, Nan Duan, Tao Qin, Zhao Yan, and Ming Zhou. 2017.
\newblock Question answering and question generation as dual tasks.
\newblock \emph{arXiv preprint arXiv:1706.02027}.

\bibitem[{Tu et~al.(2020)Tu, Huang, He, and Zhou}]{tu2020graph}
Ming Tu, Jing Huang, Xiaodong He, and Bowen Zhou. 2020.
\newblock Graph sequential network for reasoning over sequences.
\newblock \emph{arXiv preprint arXiv:2004.02001}.

\bibitem[{Tu et~al.(2019)Tu, Huang, Wang, Huang, He, and Zhou}]{tu2019select}
Ming Tu, Kevin Huang, Guangtao Wang, Jing Huang, Xiaodong He, and Bowen Zhou.
  2019.
\newblock Select, answer and explain: Interpretable multi-hop reading
  comprehension over multiple documents.
\newblock \emph{arXiv preprint arXiv:1911.00484}.

\bibitem[{Vaswani et~al.(2017)Vaswani, Shazeer, Parmar, Uszkoreit, Jones,
  Gomez, Kaiser, and Polosukhin}]{vaswani2017attention}
Ashish Vaswani, Noam Shazeer, Niki Parmar, Jakob Uszkoreit, Llion Jones,
  Aidan~N Gomez, {\L}ukasz Kaiser, and Illia Polosukhin. 2017.
\newblock Attention is all you need.
\newblock In \emph{Advances in Neural Information Processing Systems}.

\bibitem[{Welbl et~al.(2018)Welbl, Stenetorp, and
  Riedel}]{welbl2018constructing}
Johannes Welbl, Pontus Stenetorp, and Sebastian Riedel. 2018.
\newblock Constructing datasets for multi-hop reading comprehension across
  documents.
\newblock \emph{Transactions of the Association for Computational Linguistics}.

\bibitem[{Williams and Zipser(1989)}]{williams1989tf}
Ronald~J. Williams and David Zipser. 1989.
\newblock A learning algorithm for continually running fully recurrent neural
  networks.
\newblock \emph{Neural Computation}, 1(2):270--280.

\bibitem[{Wu et~al.(2016)Wu, Schuster, Chen, Le, Norouzi, Macherey, Krikun,
  Cao, Gao, Macherey et~al.}]{wu2016google}
Yonghui Wu, Mike Schuster, Zhifeng Chen, Quoc~V Le, Mohammad Norouzi, Wolfgang
  Macherey, Maxim Krikun, Yuan Cao, Qin Gao, Klaus Macherey, et~al. 2016.
\newblock Google's neural machine translation system: {{Bridging}} the gap
  between human and machine translation.
\newblock \emph{Computing Research Repository}, arXiv:1609.08144.

\bibitem[{Xu et~al.(2018)Xu, Wu, Wang, Feng, Witbrock, and
  Sheinin}]{xu2018graph2seq}
Kun Xu, Lingfei Wu, Zhiguo Wang, Yansong Feng, Michael Witbrock, and Vadim
  Sheinin. 2018.
\newblock Graph2seq: Graph to sequence learning with attention-based neural
  networks.
\newblock \emph{arXiv preprint arXiv:1804.00823}.

\bibitem[{Yang et~al.(2018)Yang, Qi, Zhang, Bengio, Cohen, Salakhutdinov, and
  Manning}]{yang2018hotpotqa}
Zhilin Yang, Peng Qi, Saizheng Zhang, Yoshua Bengio, William Cohen, Ruslan
  Salakhutdinov, and Christopher~D Manning. 2018.
\newblock {{HotpotQA}}: {{A}} dataset for diverse, explainable multi-hop
  question answering.
\newblock In \emph{Proceedings of the 2018 Conference on Empirical Methods in
  Natural Language Processing}.

\bibitem[{Yu et~al.(2020)Yu, Quan, Su, and Yin}]{Yu:20}
Jianxing Yu, Xiaojun Quan, Qinliang Su, and Jian Yin. 2020.
\newblock Generating multi-hop reasoning questions to improve machine reading
  comprehension.
\newblock In \emph{Proceedings of The Web Conference 2020}, WWW ’20, page
  281–291, New York, NY, USA. Association for Computing Machinery.

\bibitem[{Zhang et~al.(2018)Zhang, Dai, Kozareva, Smola, and
  Song}]{zhang2017variational}
Yuyu Zhang, Hanjun Dai, Zornitsa Kozareva, Alexander~J Smola, and Le~Song.
  2018.
\newblock Variational reasoning for question answering with knowledge graph.
\newblock In \emph{Thirty-Second {{AAAI}} Conference on Artificial
  Intelligence}.

\bibitem[{Zhao et~al.(2020)Zhao, Xiong, Rosset, Song, Bennett, and
  Tiwary}]{zhao2020transformer-xh}
Chen Zhao, Chenyan Xiong, Corby Rosset, Xia Song, Paul Bennett, and Saurabh
  Tiwary. 2020.
\newblock Transformer-xh: {{Multi}}-evidence reasoning with extra hop
  attention.
\newblock In \emph{The Eighth International Conference on Learning
  Representations ({{ICLR}} 2020)}.

\bibitem[{Zhao et~al.(2018{\natexlab{a}})Zhao, Ni, Ding, and
  Ke}]{zhao2018paragraph}
Yao Zhao, Xiaochuan Ni, Yuanyuan Ding, and Qifa Ke. 2018{\natexlab{a}}.
\newblock Paragraph-level neural question generation with maxout pointer and
  gated self-attention networks.
\newblock In \emph{Proceedings of the 2018 Conference on Empirical Methods in
  Natural Language Processing}, pages 3901--3910.

\bibitem[{Zhao et~al.(2018{\natexlab{b}})Zhao, Ni, Ding, and
  Ke}]{zhao-etal-2018-paragraph}
Yao Zhao, Xiaochuan Ni, Yuanyuan Ding, and Qifa Ke. 2018{\natexlab{b}}.
\newblock Paragraph-level neural question generation with maxout pointer and
  gated self-attention networks.
\newblock In \emph{Proceedings of the 2018 Conference on Empirical Methods in
  Natural Language Processing}.

\bibitem[{Zhou et~al.(2018)Zhou, Yang, Wei, Tan, Bao, and Zhou}]{zhou2018nqg}
Qingyu Zhou, Nan Yang, Furu Wei, Chuanqi Tan, Hangbo Bao, and Ming Zhou. 2018.
\newblock Neural question generation from text: A preliminary study.
\newblock In \emph{Natural Language Processing and Chinese Computing}. Springer
  International Publishing.

\bibitem[{Zhu et~al.(2019)Zhu, Li, Zhu, Qian, Zhang, and
  Zhou}]{zhu2019modeling}
Jie Zhu, Junhui Li, Muhua Zhu, Longhua Qian, Min Zhang, and Guodong Zhou. 2019.
\newblock Modeling graph structure in transformer for better {AMR}-to-text
  generation.
\newblock In \emph{Proceedings of the 2019 Conference on Empirical Methods in
  Natural Language Processing and the 9th International Joint Conference on
  Natural Language Processing (EMNLP-IJCNLP)}.

\end{thebibliography}
\bibliographystyle{acl_natbib}

\clearpage
\appendix

\section{Standard vs Multi-Hop QG} \label{sec:ques_complexity}

\begin{table}[t]
\small
\centering
\begin{tabular}{@{}l r r r@{}}
 \toprule
 \textbf{QG task} & \textbf{Words} & \textbf{Entities} & \textbf{Predicates} \\
 \midrule
 Standard (SQuAD) &  10.22 & 1.12 & 1.75 \\
 Multi-Hop (HotpotQA) & 15.58 & 2.34 & 2.07 \\
 \bottomrule
 \end{tabular}
\caption{Comparison of questions' properties in standard and multi-hop QG datasets. We show the average number of words, entities, and predicates per question.}
\label{tab:qg_comp2}
\end{table}

In this section, we present our results to illustrate the relative complexity of standard and multi-hop QG tasks. For this analysis, we compare three properties of expected output \emph{i.e.}\ questions: total words, named entities, and predicates, as we believe these represent the \emph{sufficient statistics} of the question. As a benchmark dataset of standard QG, we use the development set from SQuAD~\cite{rajpurkar2018know} and for multi-hop QG, we use the development set from HotpotQA. We extract named entities using Spacy and predicates using Open IE~\cite{stanovsky-etal-2018-supervised}. 

From the results in Table~\ref{tab:qg_comp2}, we see that multi-hop questions are almost 1.5 times longer than standard ones and also contain twice the number of entities. These results suggest that in multi-hop QG the decoder needs to generate longer sequences containing more entity-specific information making it considerably more challenging than standard QG. We also observe that multi-hop questions contain roughly 2 predicates in 15 words while standard questions contain 1.75 predicates in 10 words--- highlighting that there are fewer predicates per word in multi-hop questions compared with standard ones. This highlights that information is more densely packed within the multi-hop question as they are not expected to contain latent (or bridge) entity information.

\section{Training Details} \label{sec:train_det}
We mostly follow the model training details as outlined in~\cite{sachan-neubig-2018-parameter}, which we also describe here for convenience. The word embedding layer is initialized according to the Gaussian distribution $\mathcal{N}(0,{d}^{-1/2})$, while other model parameters are initialized using \emph{LeCun uniform initialization}~\cite{lecun1998efficient}.  
For optimization, we use Adam~\citep{kingma2014adam} with $\beta_1=0.9$, $\beta_2=0.997$, $\epsilon=1e^{-9}$. 
The learning rate is scheduled as:
\begin{math}
2 \textit{d}^{-0.5} \mathrm{min}\left(\textit{step}^{-0.5},\textit{step}\cdot 16000^{-1.5}\right).
\end{math}
During training, the mini-batch contains $12,000$ source and target tokens. For regularization, we use label smoothing (with $\epsilon=0.1$)~\cite{pereyra2017regularizing} and apply dropout (with $p=0.1$)~\cite{srivastava2014dropout} to the word embeddings, attention coefficients, ReLU activation, and to the output of each sublayer before the residual connection. For decoding, we use beam search with width $5$ and length normalization following~\cite{wu2016google} with $\alpha=1$. We also use $\lambda=0.5$ when performing joint NLL and contrastive training. 

\end{document}